\documentclass{article}

\usepackage[preprint]{neurips_2025}

\usepackage[utf8]{inputenc} % allow utf-8 input
\usepackage[T1]{fontenc}    % use 8-bit T1 fonts
\usepackage[colorlinks=true, linkcolor=black, citecolor=black, urlcolor=blue]{hyperref}       % hyperlinks
\usepackage{url}            % simple URL typesetting
\usepackage{booktabs}       % professional-quality tables
\usepackage{amsfonts}       % blackboard math symbols
\usepackage{nicefrac}       % compact symbols for 1/2, etc.
\usepackage{microtype}      % microtypography
\usepackage{xcolor}         % colors

\usepackage{algorithm}
\usepackage{algorithmic}
\usepackage{multirow}
\usepackage{graphicx} 
\usepackage{amsmath}
\usepackage{comment}
\usepackage{wrapfig}

\title{Beyond Introspection: Reinforcing Thinking via Externalist Behavioral Feedback}

\author{
  Diji Yang$^1$$^*$ 
  \qquad Linda Zeng$^2$\thanks{Equal contribution.} ~\thanks{Work done during internship at UCSC.} 
  \qquad Kezhen Chen$^3$ 
  \thanks{Work done as external collaboration.} 
  \qquad \textbf{Yi Zhang}$^1$ \\
  $^1$University of California, Santa Cruz \\
  $^2$The Harker School, 
  $^3$Meta \\
  \texttt{\{dyang39, yiz\}@ucsc.edu} \\
  \texttt{\{lindazeng979, kzchen0204\}@gmail.com} \\
}

\begin{document}

\maketitle

\begin{abstract}
While inference-time thinking allows Large Language Models (LLMs) to address complex problems, the extended thinking process can be unreliable or inconsistent because of the model's probabilistic nature, especially near its knowledge boundaries.
Existing approaches attempt to mitigate this by having the model critique its own reasoning to make corrections. However, such self-critique inherits the same biases of the original output, known as the introspection illusion.
Moving beyond such introspection and inspired by core methodologies in ethology, we propose an externalist three-step framework \textbf{D}istillation-\textbf{R}einforcement-\textbf{R}easoning (\textbf{DRR}). Rather than relying on a model's introspection, DRR evaluates its observable behaviors to provide corrective feedback.  DRR first distills the reasoner's behavioral traces, then trains a lightweight, external Discriminative Model (DM). At inference time, this DM acts as a critic, identifying and rejecting suspicious reasoning steps. This external feedback compels the LLM to discard flawed pathways and explore alternatives, thereby enhancing reasoning quality without altering the base model.
Experiments on multiple reasoning benchmarks show that our framework significantly outperforms prominent self-critique methods. Benefiting from a lightweight and annotation-free design, DRR offers a scalable and adaptable solution for improving the reliability of reasoning in a wide range of LLMs.
\end{abstract}

\section{Introduction}
Articulating a thinking process, often termed chain-of-thought, shows great success in enhancing the reasoning capabilities of Large Language Models (LLMs)~\cite{wei2022chain,openai2024o1}.
However, the extended token sequences required for such thinking are vulnerable to error accumulation~\cite{bender2021dangers,press2023measuring}.
This vulnerability stems from the probabilistic nature of autoregressive models, which can produce divergent outputs when reasoning over uncertain content (i.e., near the knowledge boundary)~\cite{huang2025survey}.

A seemingly straightforward approach to this problem would be to expand the model's knowledge boundaries through further training. However, this strategy is fundamentally limited. Knowledge boundaries are an inherent property of any model and intelligence; for any given state of training, a new frontier of uncertainty will inevitably emerge when faced with novel problems. Therefore, simply fine-tuning on more data merely shifts the boundary rather than eliminating the core issue of unreliability at its edge.

This insight suggests that a more robust solution lies not in attempting to create an omniscient model, but in equipping it with the ability to accurately assess the quality of its reasoning outputs at inference time, i.e., knowing what is unknown. The critical capability, therefore, is to reliably identify moments of uncertainty as they occur.
Building on this concept, recent research attempted self-critique mechanisms, enabling models to assess and refine their outputs by leveraging their pre-trained knowledge or by estimating confidence scores~\cite{pan2023automatically}. 
However, these strategies have limitations, linked to the introspection illusion~\cite{pronin2009introspection}, due to the models’ inherent biases and tendencies towards hallucinations~\cite{stechly2023gpt, huanglarge, stechly2024self}. Moreover, instead of providing faithful representations of internal states, LLMs often perform mere mimicry or role-playing of human introspective data~\cite{comsa2025does}.
Additionally, methods that rely on accessing internal model states are impractical for closed-source LLMs, restricting their applicability.

Inspired by core methodologies in ethology, where behavior is analyzed externally without inferring unmeasurable internal states~\cite{tinbergen1963aims, rahwan2019machine}, we propose an externalist approach to augmenting LLM reasoning. 
We introduce Distillation-Reinforcement-Reasoning (DRR), a framework that treats the LLM as a fixed reasoner and uses an external module to empower the LLM with a multi-step self-correction mechanism through an in-context reinforcement learning process~\cite{monea2024llms}.
DRR first distills the reasoner's natural problem-solving attempts into behavioral traces. It then uses this data to train a lightweight, external Discriminative Model (DM) that learns to identify patterns of successful versus flawed reasoning. At inference time, this DM acts as a critic, assessing the LLM's reasoning steps and rejecting those it deems suspicious.
This explicit external verbal feedback reward~\cite{shinn2024reflexion} compels the LLM to abandon flawed pathways and explore alternative lines of reasoning—a crucial mechanism for overcoming the cognitive fixation that often limits self-critique.
The resulting interaction forms a feedback loop analogous to reinforcement in behaviorism, where external feedback guides an agent to improve its decision-making over time. This dynamic fosters a more robust and deliberative thinking process, thereby enhancing the quality and reliability of the final answer without altering the base model's weights. Moreover, the entire procedure is supported by an automated data-generation pipeline, eliminating the need for expensive, human-labeled process supervision data.

Our contributions can be summarized as follows:

\setcounter{footnote}{0}

\begin{itemize}
\item We introduce and formalize an externalist paradigm for reinforcing LLM reasoning at inference time. The DRR framework serves as a concrete realization of this paradigm, designed to overcome the fundamental limitations and inherited biases of internal self-critique.

\item We demonstrate a model-agnostic design for this paradigm, while its automated data distillation process eliminates the need for expensive manual annotation, ensuring practical scalability.

\item We provide extensive empirical validation of the externalist approach, which is shown to be beneficial for both open-source and closed-source LLMs across various benchmarks.~\footnote{All works are open sourced at \url{https://github.com/dyang39/DRR}}
\end{itemize}

\begin{comment}    
\begin{figure}[ht]
    \includegraphics[width=0.97\linewidth]
    {figures/thinking-example.pdf}
    \caption{Inference time thinking example with multi-round Answer (A) and Rationale (R) generation. With the help of verbal feedback (ENV), the LLM refines its previous answers by applying different reasoning chains.}
    \label{fig:example}
\end{figure}
\end{comment}

\section{Related Work}
The probabilistic nature of LLMs, which are optimized for statistical plausibility over factual accuracy, is a fundamental source of potential unreliability~\cite{bender2021dangers}. Stochasticity introduced during decoding is amplified at the model's knowledge boundaries, where output distributions flatten. During extended generation, such as inference-time thinking, this allows plausible-sounding errors to propagate and compound with each subsequent token. This phenomenon of compounding errors, identified as the compositionality gap, can undermine the logical integrity of a long chain of thought~\cite{press2023measuring,lanham2023measuring}. Consequently, error accumulation can lead to a final output that is logically flawed despite appearing coherent~\cite{bender2021dangers}.
The straightforward way to combat this problem is to apply intervention on the model's reasoning chain directly after discovering the flaw, but discovering the flaw itself is challenging, as it requires the system to know what it does not know, which is a capability often referred to as meta-cognition~\cite{flavell1979metacognition,liuright} or self-awareness through self-criticism~\cite{pan2023automatically}.
Existing research on system-level self-criticism can be categorized into two approaches: feedback generated by the LLM itself (LLM Self-critics) and feedback provided by additional components within the AI system (External Feedback).~\footnote{In this work, while we explicitly distinguish between LLM Self-critics and External Feedback, we note that from the system’s perspective, all critic and feedback (whether from the LLM or external modules) remains internal to the AI system.}

    \begin{figure*}
        \centering
        \includegraphics[width=\linewidth]
        {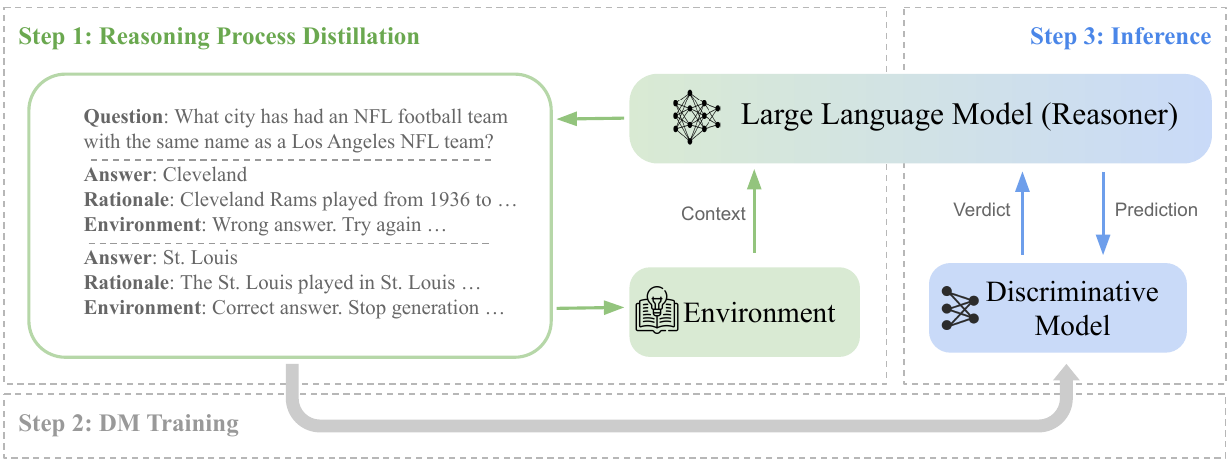}
        \caption{Overview of the three-step Distllation-Reinforcement-Reasoning (DRR) pipeline.}
        \label{fig:overview}
    \end{figure*}
    
\paragraph{LLM Self-critics}
LLM self-critic mechanisms empower language models to assess and refine their own outputs~\cite{kadavath2022language, yao2022react, asai2023self, amayuelas2023knowledge}. There are two predominant strategies for LLMs to self-present their feedback~\cite{pan2023automatically}. 
A straightforward approach leverages the model's pre-trained knowledge to critique and enhance its responses~\cite{wangself, weng2023large, madaan2024self,wang2023self, yin2023large}. While this method utilizes the model's inherent capabilities, it may not be able to reliably identify its own errors due to its inherent biases and tendencies to hallucinations~\cite{stechly2023gpt}.
The second approach involves using confidence scores or uncertainty estimates to evaluate the quality of the model’s outputs~\cite{xie2024self, farquhar2024detecting, cheninside, zhang2023enhancing}. This strategy typically requires manually crafted criteria, which can be labor-intensive and lack a theoretical foundation for optimal configuration, thereby limiting its reliability and generalizability~\cite{pan2023automatically}. Moreover, the need to access the internal state of the model closes off the opportunity for its application to good-performing closed-source LLM APIs such as ChatGPT and Gemini.

Inspired by the fact that ethologists draw conclusions about animals through observable behavior and avoid making assumptions about internal states that can hardly be directly measured~\cite{tinbergen1963aims,rahwan2019machine}, our approach emphasizes the analysis of real observable behavior (i.e., the outputs of LLM) rather than defining internal states (e.g., confidence thresholds).

\paragraph{External Feedback}
Recent studies have shown that external modules can provide meaningful feedback during the generation process of LLMs, thereby enhancing system performance~\cite{goucritic}. These modules encompass a diverse range of variations, including tools like Code Interpreters~\cite{chenteaching}, Search Engines~\cite{trivedi2023interleaving}, and other specialized software~\cite{kim2024language}, as well as carefully designed rules~\cite{yao2024tree}. MemPrompt~\cite{madaan2022memory} predefines a dictionary of possible LLM outputs and their corresponding scores to offer feedback at inference time, similar to our analysis of LLM behavior. While these approaches effectively respond to LLM behavior across various benchmarks, however, these non-deep learning tools tend to limit the flexibility of the system.

On the other hand, since RLHF~\cite{ouyang2022training}, using external deep learning models as a reward provider has also been widely studied recently~\cite{bai2022training,glaese2022improving}. These works focus on using the feedback provided by the reward model to fine-tune the LLM instead of directly participating in the generation at inference time. Consequently, these resource-intensive approaches require large amounts of high-quality labeled data to train the reward models and substantial computational resources to perform LLM tuning.

In contrast, our training only involved a small classifier fine-tuning on the LLM’s behavioral data. This data can be easily generated from any raw data with only the required initial input and final output, which are typically available for most datasets.

\paragraph{Hallucination Detection}
Hallucination detection has emerged as a pivotal task in the era of large language models, aiming to differentiate between hallucinated and accurate content using binary classification~\cite{ji2023survey}. While our work does not specifically focus on hallucination detection, it shares certain methodological similarities with one prominent approach in that field: training hallucination classifiers to detect hallucinations based on the model’s internal states, such as hidden layer representations~\cite{azaria2023internal, su2024unsupervised}. 
However, this method requires direct access to the model’s internal workings, making it unsuitable for closed-source systems.
A recent approach, RelD~\cite{chen2023hallucination}, utilizes hallucination data from LLMs and designs complex training objectives to align a hallucination detector with human evaluation metrics. 
Inspired by this line of research, our approach circumvents the complexities of human alignment and intricate training designs. 
Instead, we rely entirely on behavioral data from model outputs, which allows our DM to prevent hallucinations. More importantly, since our focus extends beyond hallucination detection, our system considers this as a mid-step signal and analyzes how this signal can be used to further optimize the reasoning chains of LLMs, thereby improving the overall performance in general natural language generation tasks.

\section{Approach}

In this section, we introduce our main framework for enhancing LLM inference-time reasoning. As shown in Figure~\ref{fig:overview}, it consists of three steps: behavioral data generation via Reasoning Process Distillation, where LLM reasoning patterns are collected as training data (Section~\ref{sec:data_generation}); training of the DM, where the DM learns to assess LLM behavior (Section~\ref{sec:training}); and system deployment for inference, during which the LLM and the DM interact over multiple iterations in order to produce a final answer or abstention (Section~\ref{sec:inference}).

\subsection{Reasoning Process Distillation} %Behavorial Data Generation
\label{sec:data_generation}

\begin{algorithm}[h]
\caption{Reasoning Process Distillation Algorithm}
\label{alg:data_generation}
\begin{algorithmic}[1]
\STATE \textbf{Notation:}
\STATE $\mathcal{R}$: Reasoner LLM
\STATE $Q$: Question, $A$: Ground-truth Answer, $C$: Context (initialized as empty)
\STATE $r$: Generated rationale, $A'$: LLM-generated answer

\FOR{each data point}
    \STATE $C \gets \emptyset$
    \WHILE{true}
        \STATE $(A', r) \gets \mathcal{R}(Q, C)$
        \IF{$A' = A$}  \label{line:verdict}
            \STATE Record $\{Q, C, A', r\}$ with verdict \texttt{Accept}
            \STATE \textbf{break} \COMMENT{Exit the while loop}
        \ELSE
            \STATE Record $\{Q, C, A', r\}$ with verdict \texttt{Reject}
        \ENDIF
            \STATE $C \gets C \cup \{A', r\}$

        \STATE \textbf{exit} while loop \textbf{if} maximum turns reached.
    \ENDWHILE
\ENDFOR

\end{algorithmic}
\end{algorithm}

Algorithm~\ref{alg:data_generation} depicts our semi-supervised Reasoning Process Distillation algorithm, which collects data for DM training. 
This algorithm simulates the LLM’s inference-time behavior in multiple reasoning iterations using raw training data.
For each question $Q$, the Reasoner LLM $\mathcal{R}$ generates an answer $A'$ along with its rationale $r$. A binary verdict label is assigned to indicate whether the DM should accept or reject the response, depending on whether $A'$ matches the ground-truth answer $A$.
For all incorrect answers, $\mathcal{R}$ is re-prompted with the question $Q$ and past history (i.e., context $C$) to explore alternative reasoning paths. 
This iterative process continues until $\mathcal{R}$ generates the correct answer $A$, signaling that the DM should accept the response and stop further reasoning. If no accurate response is generated after a predefined number of iterations, the process terminates to avoid an infinite loop (more discussions in Appendix~\ref{appendix:max_turns}).
The generated data reflect a diverse range of LLM behaviors that the DM may encounter across various turns during inference, allowing the DM to directly learn the unconscious reasoning patterns that LLM tends to generate when leading to a correct or incorrect answer.
By leveraging LLM's natural behaviors with its environment, this generation method streamlines the feedback collection process, relying solely on the input and label from source data, which are typically available for most NLP and AI datasets.

\subsection{Discriminative Model Training} 
\label{sec:training}

Given a question, its past context, and an LLM-generated answer with rationale, the goal of the DM is to predict whether the new answer should be accepted or rejected. In other words, the DM acts as a binary classification model, and its training on a
distribution over \(\bigl(x,y\bigr)\)
is alignment between the input tuple $x_i = \{Q, C, A', r\}$ and the label $y \in \{\texttt{Accept}, \texttt{Reject}\}$. Through observing both the LLM's previous responses and its new rationale, which may include corrections to past reasoning or further explorations into internal knowledge, the DM learns to be both a coach, supervising the LLM's behavior across multiple iterations, and a judge, determining the reliability of its final response. This enables the DM to determine if the LLM’s response is influenced by hallucinations and when it should stop.

\begin{comment}
Specifically, the DM predicts logits $\mathbf{z} \in \mathbb{R}^2$, representing unnormalized probabilities for the two classes. 
The training objective is to minimize a cross-entropy loss as shown in Equation~\ref{eq:loss}, where $P(y|x;\theta)$ is the predicted probability of the class $y$.
\begin{equation}
\mathcal{L}_{\text{DM}}(\theta) = - \,\mathbb{E}_{(x,y)\,\sim\,\mathcal{D}} \bigl[\log P\bigl(y \mid x;\theta\bigr)\bigr] 
\label{eq:loss}
\end{equation}
% To develop the DM model, we train a language model for binary classification using the generated data from Section~\ref{sec:data_generation}. Given a question, its context, including past search queries and retrieved documents, the LLM's answer, and the LLM's rationale, the DM model predicts if the answer should be accepted or rejected. It relies on the DM model to determine if the LLM's answer occurs due to hallucinations or not. 
% Weighted training to be discussed
% Refer to 4.2 for implementation details
% Refer to 6 for the impact of false positive
Using the behavioral data described in Section~\ref{sec:data_generation}, the DM learns diverse scenarios reflective of inference-time conditions. Each reasoning turn is a data point, and all turns for a question remain in the same set to prevent knowledge leakage. The usage of behavioral data eliminates reliance on external reward signals, commonly collected through human feedback, as the DM inherently models these features by learning directly from the dataset.
\end{comment}

Specifically, the DM predicts logits $\mathbf{z} \in \mathbb{R}^2$, representing unnormalized probabilities for the two classes. To account for the higher cost of false-positive errors (i.e., Accepting an incorrect answer), the weight $w_1$  for class \texttt{Accept} is set higher than $w_0$, the weight for class \texttt{Reject}. The weighting design encourages DM to adopt stricter acceptance criteria, prioritizing the reliability of the final decision and reducing the likelihood of harmful false-positive predictions (see Section~\ref{sec:fp} for details). 
The training objective is to minimize a weighted cross-entropy loss as shown in Equation~\ref{eq:loss}, where $P(y|x;\theta)$ is the predicted probability of the class $y$, and $w_{y}$ represents class-specific weights. 

\begin{equation}
\mathcal{L}_{\text{DM}}(\theta) = - \,\mathbb{E}_{(x,y)\,\sim\,\mathcal{D}} \bigl[w_{y}\,\log P\bigl(y \mid x;\theta\bigr)\bigr]
\label{eq:loss}
\end{equation}

By utilizing the behavioral data described in Section~\ref{sec:data_generation}, the DM is exposed to diverse scenarios reflective of inference-time conditions. This approach eliminates the reliance on external reward signals, commonly collected through human feedback, as the DM inherently models these features by learning directly from the dataset.

\subsection{Inference} 
\label{sec:inference}

At inference, the system features an iterative line of exchanges between the Reasoner LLM and the DM. While the LLM's goal is to provide a final answer to a given question, the DM assesses LLM reasoning and provides feedback signals through accepting or rejecting its responses.

Similar to the data generation process in Section~\ref{sec:data_generation}, the LLM produces answers and rationales across multiple iterations. However, during inference, the stopping condition is based on the DM's feedback rather than comparisons to the ground truth. The DM evaluates each LLM-generated answer and rationale in real-time, predicting \texttt{Accept} or \texttt{Reject} verdicts based on the question, context, answer, and rationale of each turn. This process guides the system toward a reliable final response through recognizing patterns in responses that indicate hallucinations or errors in reasoning.
% This feedback mechanism enables the system to operate without requiring ground-truth comparisons during inference.
In general, this architecture offers flexibility by supporting adaptive reasoning iterations. The DM dynamically decides whether further exploration is necessary based on the behavioral data it was trained on. This allows the system to efficiently balance between preventing overextended reasoning and addressing incomplete or incorrect responses. This adaptability mirrors human problem-solving, where some answers are instinctively clear, while others demand deeper reasoning.

\section{Experiment}

\begin{table*}[tb]
    \centering
    \resizebox{\textwidth}{!}{
    \renewcommand{\arraystretch}{1.18}
    \setlength{\tabcolsep}{3pt}
        \begin{tabular}{l l l c c c c c c c c c c c c c c c}
            \toprule
    
            & & & \multicolumn{6}{c}{Commonsense} & \multicolumn{6}{c}{Knowledge Intensive} & \multicolumn{3}{c}{Overall}\\
            \cmidrule(lr){4-9} \cmidrule(lr){10-15} \cmidrule(lr){16-18} 
    
            & & & \multicolumn{3}{c}{CommonSenseQA} & \multicolumn{3}{c}{WinoGrande} &
            \multicolumn{3}{c}{OpenBookQA} & \multicolumn{3}{c}{PIQA} & \multicolumn{3}{c}{Combined} \\ 
            \cmidrule(lr){4-6} \cmidrule(lr){7-9} \cmidrule(lr){10-12} \cmidrule(lr){13-15} \cmidrule(lr){16-18}
            
            & & & Acc & FS(1) & FS(3) & Acc & FS(1) & FS(3) & Acc & FS(1) & FS(3) & Acc & FS(1) & FS(3) & Acc & FS(1) & FS(3) \\
            
            \midrule
            \multirow{6}{*}{\rotatebox{90}{\color{gray}\small{Open Source}}} 
            & \multirow{3}{*}{\rotatebox{90}{\color{gray}\small{ZS}}} 
            & Self-Talk (Llama3)* & 70.6 & -&-&- & - & - & 72.2 & - &-& 77.2 & -&- & - & -&- \\
            && CoT (Llama3) & 73.5 & 47.0& -6.1 & 59.0 & 17.9&-64.2 & 70.0 & 40.0&-20.0 & 78.1 & 56.3 & 12.5 & 72.5 & 48.5& -9.9 \\
            && Abstain (Llama3) & 73.1 & 46.3&-7.5 & 59.3 & 18.9&-62.9 & 70.2 & 40.4&-19.2 & 66.5 & 33.1&-33.8 & 66.5 & 35.6& -33.4 \\
            
            \cmidrule(lr){2-18}
            & \multirow{3}{*}{\rotatebox{90}{\color{gray}\small{Learned}}} 
            & Crystal (Llama3)* & 75.1 & -&-&- & - & - & 72.6 & -&- & 78.2 & - &-& - & -&- \\
            &&  SFT-LoRA (Llama3) & 77.2 & 54.5 & 8.9 & 76.4 & 52.8 & 5.6 & 80.6 & 61.2 & 22.4 & 82.0 & 64.0 & 28.0 & 79.8 & 54.5 & 19.0 \\
            && DRR (Llama3) & \textbf{82.3} & \textbf{65.5}& \textbf{33.6} & \textbf{77.0} &\textbf{54.1}&\textbf{8.1} & \textbf{81.2} & \textbf{63.0}& \textbf{26.6} & \textbf{83.6} & \textbf{67.3}& \textbf{34.8} & \textbf{81.8} & \textbf{65.5}& \textbf{28.3} \\
            
            \midrule
            \multicolumn{2}{c}{\multirow{4}{*}{\rotatebox{90}{\color{gray}\small{Closed Source}}}}
            & CoT (GPT-4) & 83.5 & 67.1 & 34.2 & 74.6 & 49.2 & -1.7 & \textbf{91.2} & 82.4  & 64.8 & 91.1 & 82.2 & 64.3 & 86.1 & 65.5 & 44.5 \\
            && Abstain (GPT-4) & 82.6 & 65.1 & 30.2 & 74.4 & 48.9 & -2.3 & 85.6 & 71.2 & 42.4 & 85.4 & 70.8 & 33.0 & 82.6 & 59.3 & 25.8 \\
            && Self-Critic (GPT-4) & 74.6 & 49.2& -1.6 & 77.0 & 54.1 & 8.1 & 88.8 & 77.6 & 55.2 & 88.7 & 77.4 & 54.7 & 83.4 & 60.5 & 33.7 \\
            && DRR (GPT-4) & \textbf{86.2} & \textbf{74.4}&\textbf{52.2} & \textbf{79.5} & \textbf{59.2}&\textbf{19.9} & \textbf{91.2} &\textbf{82.8}& \textbf{67.6} & \textbf{91.5} & \textbf{82.9}& \textbf{66.3} & \textbf{87.9} & \textbf{69.4}& \textbf{53.9} \\

            \bottomrule
        \end{tabular}
    }
    \caption[Results on four question answering datasets]{Results on four question answering datasets. Each method is evaluated by Accuracy (Acc) and two Formula Scores FS(1) and FS (3). The Overall (Combined) results aggregate all data points across datasets into one unified set for holistic evaluation. Methods marked with * indicate results reported by prior works.\setcounter{footnote}{2}\footnotemark}
    \label{tab:result}
\end{table*}

\subsection{Environment Setup}
\paragraph{Tasks and Datasets} 

Our approach is evaluated on commonsense and knowledge-intensive reasoning tasks. The former involves general, everyday scenarios, while the latter requires applying specialized factual knowledge to solve complex problems. We assess commonsense reasoning with CommonsenseQA \cite{talmor2019commonsenseqa} and WinoGrande \cite{sakaguchi2020winogrande}, and knowledge-intensive reasoning with OpenBookQA \cite{mihaylov2018can} and PIQA \cite{bisk2020piqa}. To generate mid-step behavioral data for DM training, we use QA pairs from the training sets. We evaluate OpenBookQA on the test set and the other datasets on the validation sets.

\paragraph{Evaluation}\label{sec:evaluation}

We use standard exact matching accuracy (Acc) as our primary measure; however, it may not fully capture user satisfaction, since an incorrect or misleading answer can be more disappointing than no answer. 
Therefore, we also include Formula Score (FS)~\cite{davis1967note}, inspired by scoring systems in mathematical competitions~\cite{AMC_10}, which assign negative scores to incorrect answers and no penalties for abstentions. 
Specifically, each correct answer yields $S_{\text{correct}}$, each incorrect answer yields $S_{\text{incorrect}}$, and an abstention yields $S_{\text{abstain}}$. 
n our experiments, these are set to $1$, $-1$, and $0$, respectively, which we denote as FS(1). This scoring scheme can be generalized to adjust the relative penalties, depending on the importance of reliability. For instance, FS(3) increases the penalty for errors to $-3$, placing greater emphasis on avoiding incorrect answers. By reporting both accuracy and FS, we provide a more balanced evaluation of QA performance, especially in settings where reliability and safety are critical (e.g., medical or legal advice).

Additionally, we compute the critic-decision accuracy Acc(D), which measures the accuracy of a system's decisions to give an answer or abstain (i.e. how well it decides to abstain or not). For DRR, this refers to the DM accuracy at the last reasoning turn before an answer is given. 
For zero-shot methods, this refers to the percentage of LLM's successful answers and abstentions.

\subsection{Implementation Details}

\paragraph{Model Choice}
We use Llama3-8b~\cite{dubey2024llama} and GPT-4 as the LLM, and we finetune the Flan-T5-783M~\cite{chung2024scaling} as the DM. Our choice of LLMs reflects both open-source and closed-source models to demonstrate the flexibility of our system.
Although the DM can take on any classifier architecture, we adopt Flan-T5 for its strong trade-off between performance and size on binary classification tasks. Smaller discriminative models are particularly practical, as they require fewer parameters, retrain efficiently, and operate independently of the main LLM. Prior work, such as Tyen et al. \cite{tyen2023llms}, has also shown the effectiveness of lightweight classifiers for detecting LLM errors. Building on these insights, our design leverages the LLM’s generative capacity and the DM’s discriminative specialization to form an efficient self-correction framework, reducing the training burden on the LLM while constructing a simple classification task that a small DM can perform. 

\paragraph{Fine-tuning Details}
The DM is fine-tuned on the behavioral data generated for all datasets from the first Reasoning Process Distillation step of DRR. Note that our method does not require any optimization for individual datasets; that is, for all experiments, we train one classifier for all four datasets.
To account for the more serious consequences of false-positive errors (i.e., accepting an incorrect answer; see Section~\ref{sec:fp} for more explanation), the weight factor $w_1$  for class \texttt{Accept} is set three times higher than $w_0$, the weight for class \texttt{Reject}. During loss calculation, this class-specific weight factor is multiplied by the cross-entropy shown in Equation~\ref{eq:loss}.
For reproduction concern, we detail the data preparation process, training hyperparameters, prompt configurations, and computing cost in Appendices~\ref{appendix:training} and~\ref{appendix:prompts}.

\footnotetext{The original works use models different from Llama3. For a fair comparison, we report the performance from recent reproduction~\cite{molfese2024zebra}. Due to limited access to the prediction file, the FS and Overall score are unavailable.}

\section{Results and Analysis}

\paragraph{Baseline methods}
Table~\ref{tab:result} categorizes baselines by model availability, distinguishing between Open Source and Closed Source methods.

For baselines using an open-source model, we first include three zero-shot methods. 
Self-Talk~\citep{shwartz2020unsupervised}, as one of the pioneering works leveraging model self-awareness (for information seeking) in a zero-shot setting, has recently been enhanced with the support of a LLM (Llama3) and updated prompting methods, establishing it as a strong baseline~\cite{molfese2024zebra}. 
Then, we evaluate two widely used prompting-based methods: CoT used a well-tuned QA prompt to encourage LLM to think step by step, while Abstain included an abstain~\cite{wen2024know} option to give up answering when LLM thinks it is not confident.
Notably, DRR includes abstentions by treating any instances where the DM consistently rejects responses until the maximum number of attempts is reached as an abstention action.
In addition to ZS approaches, we evaluate two Learned methods.
Crystal~\cite{liu2023crystal} is a system-level self-critic approach similar to Self-Talk, but it tunes the model for self-feedback ability via reinforcement learning. 
% While the original work uses T5 as the Reasoner backbone, we report results from a recent Llama3 reproduction~\cite{molfese2024zebra} to ensure model consistency in comparisons.
We also conduct a supervised fine-tuning baseline (SFT-LoRA) to enable a more direct and fair comparison to our approach using the same amount of training data. To match the trained parameters used in our system, we apply Low-Rank Adaptation (LoRA)~\citep{hulora} for parameter-efficient fine-tuning. This baseline highlights the comparison between the standard outcome fine-tuning approach and DRR.

For methods supporting Closed Source models, we include CoT and Abstain under the same settings as above but with strong backbone LLM (GPT-4). Additionally, we built a prompt-based self-reasoning baseline (Self-Critic) following the implementation by Huang et al.~\cite{huanglarge}. This evaluates the LLM’s self-awareness to refine its responses without external feedback.
 
\paragraph{Results} %conclusion first: we outperform all baselines + methods in open and also closed. Individual method comparison - use conclusion + report overall number.
Table~\ref{tab:result} reports the experimental results. DRR outperforms all baselines under both Open Source and Closed Source settings.

For the Open Source setting, DRR with Llama3 demonstrates consistent improvements in all metrics from the three ZS baselines. Raising overall accuracy from the CoT baseline of 72.5\% to 81.8\%, DRR demonstrates its ability to enhance self-correction via iterative reasoning. While the Abstain allows the LLM to opt-out, it performs worse than CoT, even in FS(3), indicating that LLMs struggle with the self-awareness to abstain at the right time.
Compared to learned methods, DRR surpasses Crystal on three available datasets, benefiting from external DM feedback without requiring expensive RL. Unlike SFT-LoRA, which does not provide reasoning chains during inference time, DRR generates a mid-step thinking path, improving interpretability and achieving a 2.5\% accuracy gain. These gains are even more pronounced in FS, where DRR outperforms the Abstain baseline by 29.9\% for FS(1) and 61.7\% FS(3), showing its effectiveness in mitigating incorrect responses through selective abstention.

   \begin{wrapfigure}{r}{0.5\columnwidth}
        \centering
        \includegraphics[width=0.5\columnwidth]{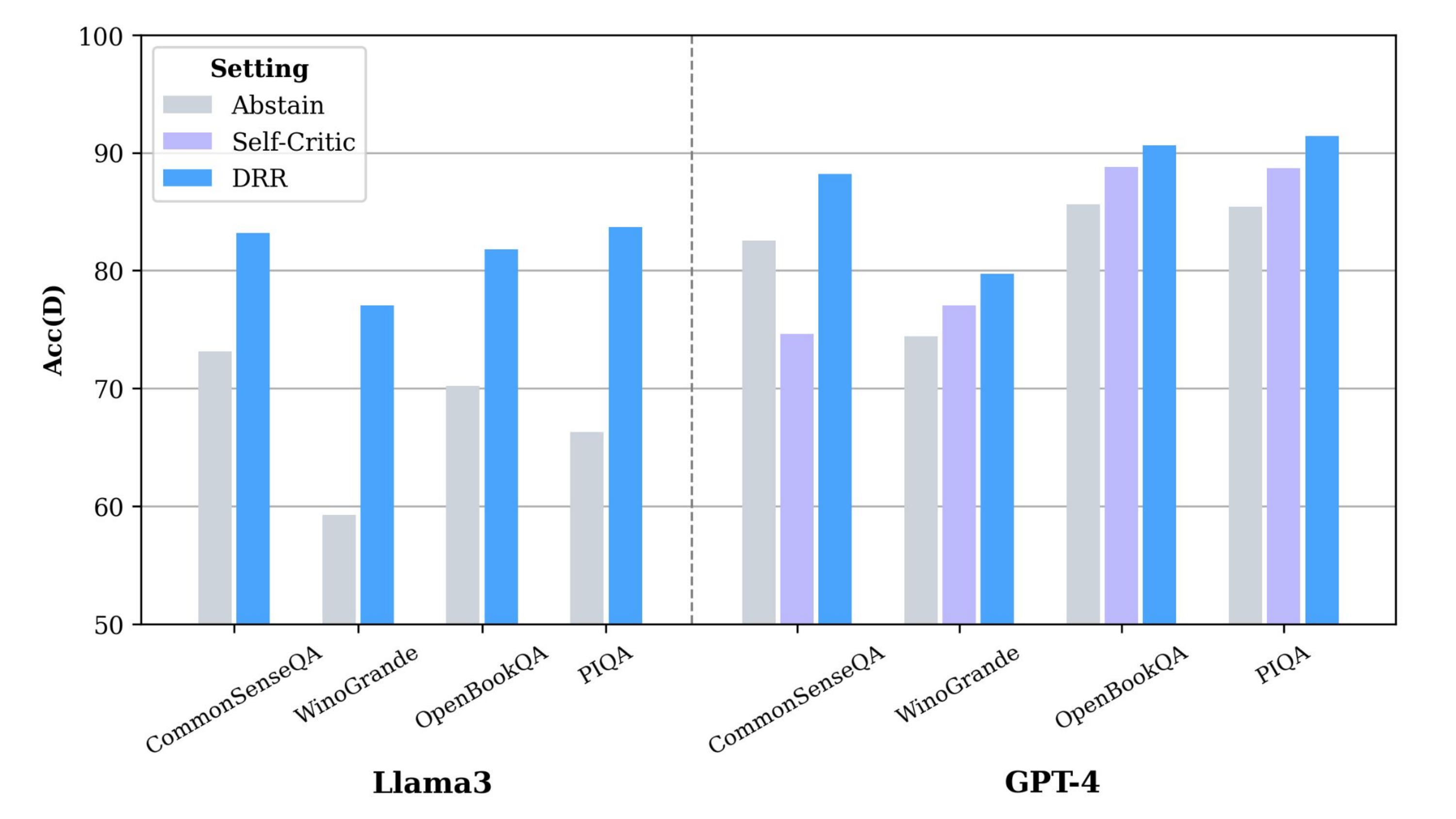}
        \caption{Critic-decision accuracy Acc(D) for Abstain, Self-Critic, and DRR settings using Llama3 and GPT-4 as Reasoner.}
        \label{fig:accd}
    \end{wrapfigure}
    
For the Closed Source setting, DRR with GPT-4 surpasses CoT, Abstain, and Self-Critic by 1.8\%, 5.3\%, and 4.5\% in overall accuracy, respectively. 
DRR's high improvement over Self-Critic in FS(3) (53.9\% vs. 33.7\%) emphasizes the advantage of external feedback over self-reasoning, especially for closed-source models where fine-tuning is infeasible.
Notably, Self-Critic performs worse than CoT, dropping 3.1\% in accuracy, aligning with Huang et al.~\cite{huanglarge} on LLM self-awareness limitations.
With heightened punishment to errors, Formula Score further illustrates DRR's advantage: while accuracy gains 5.3\% from Abstain, FS(1) and FS(3) increase by 10.1\% and 28.1\%, demonstrating its effectiveness in scenarios where reliability is critical and abstaining is preferred over hallucinating.

\paragraph{Critic-Decision Accuracy}

Figure \ref{fig:accd} compares DRR's critic-decision accuracy against Abstain and Self-Critic baselines, demonstrating the DM's positive impact on inference-time decision-making. 
As previously discussed, the critic’s decision ability inherently depends on the model’s self-awareness of its knowledge. 
Using Llama3 as the Reasoner, DRR achieves 81.4\% accuracy across four datasets, surpassing Abstain's 67.2\%.
For GPT-4, DRR reaches 87.5\%, outperforming Abstain (82.0\%) and Self-Critic (82.3\%) across all datasets.
%For GPT-4, while both Abstain (82.0\%) and Self-Critic (82.3\%) baselines exhibit much stronger capability on Acc(D) than Llama3-Abstain, DRR achieves the highest average accuracy at 87.5\%, consistently outperforming both baselines across all datasets.
By achieving better Acc(D) compared to baselines where the LLM makes decisions independently, the use of the DM facilitates more accurate answers and meaningful abstentions during inference-time reasoning.

    \begin{figure*}[!t]
        \centering
        \includegraphics[width=\linewidth]
        {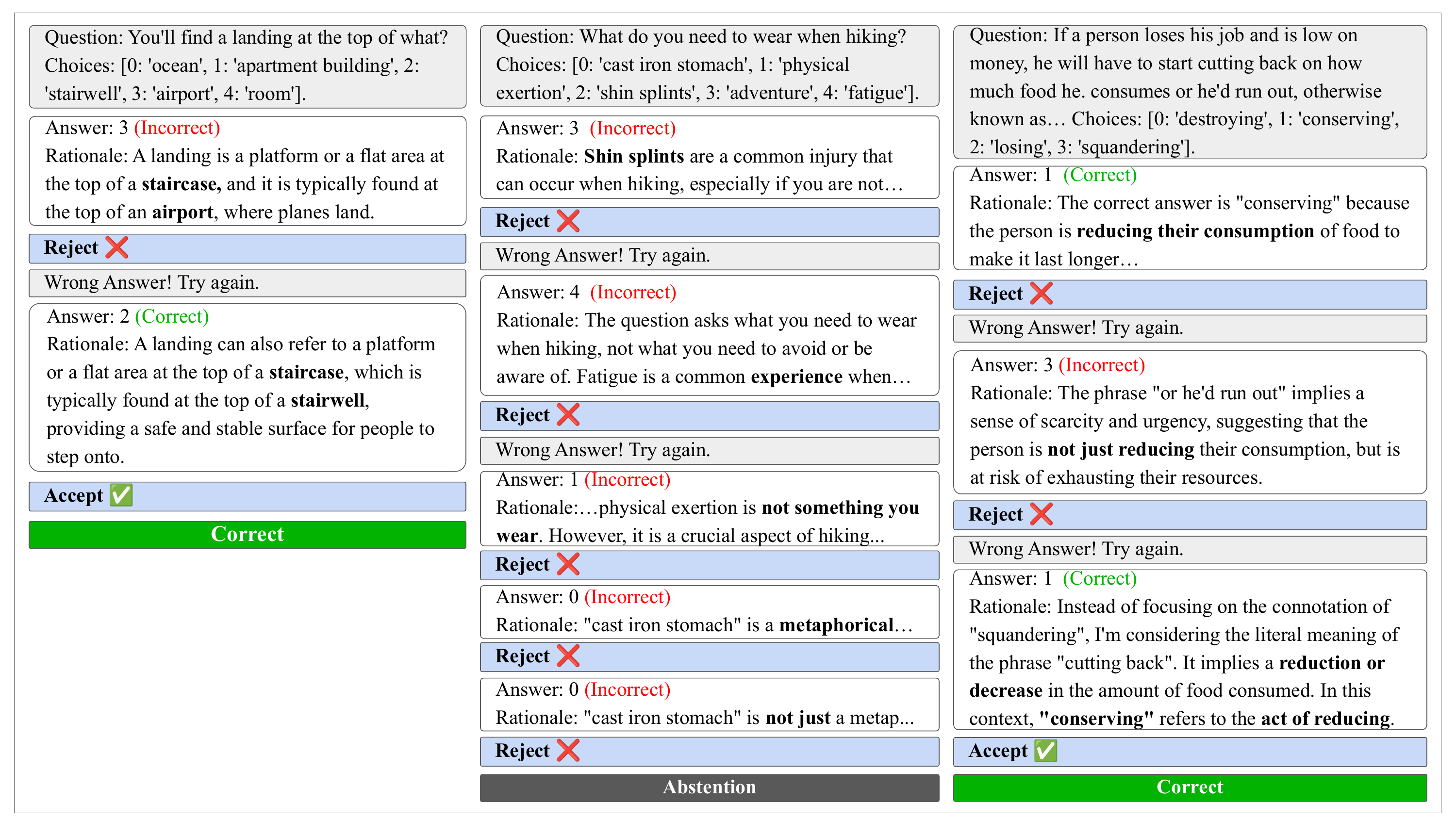}
        \caption{Examples from CommonsenseQA and OpenBookQA with DRR. \textit{Left}: successful correction and acceptance; \textit{Center}: continued rejection leading to abstention; \textit{Right}: false rejection followed by successful acceptance.}
        \label{fig:examples}
    \end{figure*}

\section{Discussion}

\paragraph{Qualitative Examples}

Figure \ref{fig:examples} displays various examples of the common behavior of our system at inference time. The first example shows optimal behavior, where the LLM responds incorrectly with disjoint reasoning in its rationale on the first turn, the DM identifies this hallucination and rejects the response, and the LLM retries on the second turn, adequately correcting its answer and rationale, leading to an accepted correct answer. We also observe this behavior occurring flexibly across more than two turns. The second example shows our system's ability to mitigate cases where LLM can never reach a correct answer. The DM consistently rejects faulty responses, where the LLM answer does not match an otherwise correct rationale (Turn~1), where the rationale contradicts itself (Turn 2), or where the rationales do not directly answer the question (subsequent turns), until the maximum turns are reached, leading to abstention. While these abstentions are considered incorrect answers for the accuracy score, they mitigate the negative effects measured by FS, highlighting the metric's importance. The third example demonstrates the system's ability to mitigate errors by correcting itself, both on the LLM and DM sides. Initially, the DM incorrectly rejects the answer, but later corrects itself and accepts the correct response. This illustrates how our system can prompt the LLM to reconsider its initial answer and provide a more logical and well-explained rationale in subsequent turns. This approach reduces the impact of false negatives from the DM, making such errors less costly to overall system performance.
%The third example shows the DM incorrectly rejecting an answer at the first turn, then correcting itself, and accepting the correct answer at a later turn. This demonstrates our system's capability to mitigate errors by correcting itself on both the LLM and DM side. In fact, it can make the LLM reconsider its initial response and be more sure about the next time around, as shown by the more logical and well-explained rationale at the later turn. Furthermore, this shows the less weight put on DM's false negatives, making them less costly to system performance.

\paragraph{Generative vs. Discriminative}

Our system relies on a lightweight DM to identify faulty reasoning steps rather than a single generative model handling both answer production and error detection. We argue that discriminative decisions, such as binary classification of correctness, can be less complex than open text generation. The DM only needs to learn whether the LLM’s output is acceptable or flawed, rather than produce the entire answer. In practice, as illustrated in Figure~\ref{fig:examples}, identifying a hallucinated statement or a gap in reasoning is often more straightforward than generating a complete solution from scratch.

This observation aligns with adversarial training research in image generation, where the discriminator often outperforms the generator in classification, sometimes leading to mode collapse~\cite{goodfellow2014generative,arjovsky2017wasserstein}. Although we differ from image-generation settings, this suggests that a specialized classifier can outpace a generative model at spotting errors. Similarly, recent self-critique work shows that LLMs are not inherently strong in discriminative tasks like self-correction unless trained for them~\cite{jiang2024self}. In our approach, the DM handles binary classification, freeing the LLM to focus on generative reasoning.

\paragraph{The Impact of False Positives}
\label{sec:fp}
% weighted training
False positives, where the DM incorrectly accepts a wrong LLM answer, are more costly compared to DM rejections in our system because acceptances immediately end the reasoning process and provide no opportunity for future correction. On the other hand, false negatives, which are rejections of correct answers, proceed to another round of reasoning, where the DM or the LLM can correct its behavior (i.e., the third example in Figure \ref{fig:examples}). Motivated by the idea that the DM should be more cautious in choosing to accept an answer than to reject one, we introduce weighted training to the DM training (detailed in Section \ref{sec:training}). This weighting design encourages DM to adopt stricter acceptance criteria, prioritizing the reliability of the final decision and reducing the number of harmful false-positive predictions.

\section{Limitations}
Inference-time latency is one of the key limitations of the DRR framework. Unlike single-pass approaches with one LLM call, such as standard Chain-of-Thought prompting, DRR requires multiple interactions between the LLM and the DM. Similar to other models with inference-time thinking capability like OpenAI-o1~\citep{openai2024o1}, the thinking period makes the method less suitable for real-time or latency-sensitive applications where efficiency is a primary concern.

To preserve pre-trained LLM's well-tuned alignments and to support the use of closed-source models for real-world usability, this paper does not explore modifying model weights or accessing internal states. Although this makes DRR broadly applicable, it also limits the system to the pre-trained capability of LLM. Training the LLM could improve both the LLM's reasoning and the quality of behavioral data for the DM, potentially enhancing inference-time reasoning. Studying DRR in trainable settings remains a meaningful direction for future work.

While DRR is designed to be model-independent and can be used in a wide range of domains, its generalizability to specialized domains such as law, medicine, or scientific reasoning remains to be explored. Due to space constraints, we leave the extensive domain adaptation studies to future work. Exploring strategies to improve robustness in different domains, such as incorporating domain-specific cues or augmenting training data, could further improve the applicability of DRR in real-world scenarios.

\section{Conclusion}
LLMs often exhibit unreliable inference-time reasoning, where self-critique fails due to inherited biases. Furthermore, LLMs face the pressing challenge of diminishing fresh training data as they need ever-increasing high-quality inputs to refine their advanced capabilities. To address these challenges, we proposed Distillation-Reinforcement-Reasoning (DRR), a framework that enhances LLM's inference-time reasoning via in-context reinforcement learning supervised by a lightweight discriminative model. This model is trained from synthetic data generated from the LLM through a novel reasoning process distillation algorithm, providing an avenue for generating fresh behavioral training data without relying on introspection. 
Empirical results on multiple standard QA benchmarks demonstrate the effectiveness of this framework over self-critique approaches.
We expect this low-cost and externalist solution to provide a practical path for not only LLM, but also more complex agentic systems~\cite{zhao2025llm} to continually improve their reasoning quality and reliability at inference time without relying on flawed introspection.

\bibliographystyle{plain} % We choose the "plain" reference style
\bibliography{refs} % Entries are in the refs.bib file

%%%%%%%%%%%%%%%%%%%%%%%%%%%%%%%%%%%%%%%%%%%%%%%%%%%%%%%%%%%%

\appendix
\newpage

\section{Training Details}\label{appendix:training}

\begin{wraptable}{r}{0.65\linewidth}
\centering
\caption{DM training and development dataset sizes across four QA benchmarks, with Llama3 and GPT-4 as the Reasoner.}
\renewcommand{\arraystretch}{1.2}
\begin{tabular}{lcccc}
\toprule
\textbf{Dataset} & \multicolumn{2}{c}{\textbf{Train}} & \multicolumn{2}{c}{\textbf{Dev}} \\
\cmidrule(lr){2-3} \cmidrule(lr){4-5}
 & Llama3 & GPT-4 & Llama3 & GPT-4 \\
\midrule
CommonSenseQA  & 11416 & 11124 & 2855 & 2783 \\
Winogrande     & 10969 & 10529 & 2744 & 2668 \\
PIQA           & 7785  & 3654  & 1950 & 911  \\
OpenBookQA     & 5496  & 4909  & 1365 & 1230 \\
\midrule
\textbf{Overall} & \textbf{35666} & \textbf{30216} & \textbf{8914} & \textbf{7592} \\
\bottomrule
\end{tabular}
\label{tab:dataset_comparison}
\end{wraptable}

 \paragraph{Data Preparation}
To prepare the raw generated data in Section \ref{sec:data_generation} for DM training, we down-sample data points with the \texttt{Reject} label while keeping all \texttt{Accept} data points constant for each dataset. We observe that generated data naturally consists of more rejections than acceptances of LLM
answers because the condition for generation to
continue is a rejection. As a result, examples of rejections
are present in many turns for a question (e.g., an answer
is repeatedly rejected), but acceptances are only present in
the final turn. When down-sampling, we directly remove data points without keeping data points of different turns of conversations with the same original question (same ID) together. We make this choice because the past history of answers provided in the task content of data points already shows the full context and does not require all turns to be in the training data. Then, for each dataset, we randomly split the generated data into a train and dev set for DM training using an 80/20 split, keeping data points of different turns with same ID together to avoid data leakage. Last, we concatenate the prepared train sets and the dev sets of each dataset together. Table \ref{tab:dataset_comparison} displays the final size of the train and dev sets for Llama3 and GPT-4 experiments, as well as the relative size of each dataset.

\paragraph{Training Setting}
Prior to training, we map \texttt{Accept} labels to 1, \texttt{Reject} labels to 0, and include a basic instruction to the DM. We fine-tune Flan-T5 with an Adam optimizer ($\eta = 3e^{-5}, \lambda = 1e^{-4}$) with 500 warmup steps. The DM is fine-tuned using the Adam optimizer with the loss function described in Equation~\ref{eq:loss}. All experiments were conducted on two NVIDIA A6000 GPUs.% We use a batch size of 4, a maximum input length of 1024 tokens, and a target length of 64 tokens. 

\paragraph{Maximum Turns}
\label{appendix:max_turns}

While our ideal system could run for any number of necessary reasoning turns, our experiments are constrained by time and computational resources, and after a large amount of iterations, there is a higher possibility of an infinite state where the LLM never generates a correct answer. Consequently, we set a maximum number of turns of four for generation and five for inference, after which wrong answers are counted as abstention during inference.

\section{Prompts}
\label{appendix:prompts}

%This section outlines our baseline prompts, our system prompt settings, and examples of the specific prompts used in our experiments.

\paragraph{Zero-Shot Baselines}
For zero-shot baselines, we designed three types of prompts: (a) \textit{Standard QA Prompt}: a straightforward question-answering prompt; (b) \textit{Abstain QA Prompt}: an extended prompt allowing the model to abstain from answering if none of the options are correct; and (c) \textit{Self-Critic Prompt}: a prompt adapted from \cite{huanglarge} where the model critiques its own response and iteratively improves it. Zero-shot predictions are generated by evaluating the first turn of our system, ignoring DM outputs. Abstain zero-shot predictions are run separately, with temperature 0.1. Note that the self-critic prompt directly from \cite{huanglarge} may produce discrepancies, such as self-critic scores being lower than zero-shot scores using our system prompts. Self-critic baselines use temperature 0, following \cite{huanglarge}.

\paragraph{DRR Prompt Settings}
DRR employs two distinct prompt strategies. The first strategy uses a direct instruction-based approach.  The first turn uses the \textit{Standard QA Prompt} as a system prefix, mimicking zero-shot performance. Subsequent turns replace the standard QA prompt with a \textit{Exploration Prompt}, instructing the LLM to explore new reasoning paths based on feedback such as "Wrong Answer! Try again." Temperature settings are 0.1 (Top-P 0.9) for the first turn to ensure consistent zero-shot responses and 0.6 (Top-P 0.7) for later turns to increase diversity. The second strategy adopts a gradual prompting approach. All turns use the \textit{Standard QA Prompt} as a system prefix. The appended context includes an environment message that emphasizes gradual improvement and encourages exploring new reasoning paths without explicitly labeling past answers as wrong. Temperature settings are 0.6 (Top-P 0.9) for all turns. For a given experiment, the same strategy is used for both the generation and inference steps. The first strategy is used for Llama models, while the second strategy is applied to GPT models.

\paragraph{Prompt Examples} Examples of each prompt are displayed as following.

\noindent\textbf{Standard QA Prompt (Llama3 Zero-Shot):}

\begin{quote}
You are a knowledgeable question-answering assistant, specializing in multiple-choice questions. Based on the question and the list of choices provided, select the best answer. Carefully evaluate each option before deciding. Provide your choice (e.g., 0, 1, 2, etc) along with a brief explanation of your reasoning.

Respond only with the following format, nothing else:
Answer: [Provide the answer here]
Rationale: [Provide the rationale here]

Do not include any additional text, headers, or explanations outside this format.
\end{quote}

\noindent\textbf{Standard QA Prompt (GPT-4 Zero-Shot):}
\begin{quote}
You are a highly knowledgeable assistant skilled in multi-step reasoning for multiple-choice question answering. Based on the question and the list of choices provided, select the best answer. Carefully evaluate each option before deciding. Provide your choice (e.g., 0, 1, 2, etc) along with a brief explanation of your reasoning.
Respond only with the following format, nothing else:
Answer: [Provide the answer here]
Rationale: [Provide the rationale here]

Do not include any additional text, headers, or explanations outside this format.
\end{quote}

\noindent\textbf{Abstain QA Prompt (Llama3/GPT-4 Zero-Shot):}
\begin{quote}
You are a highly knowledgeable assistant skilled in multi-step reasoning for multiple-choice question answering. Based on the question and the list of choices provided, select the best answer. Carefully evaluate each option before deciding. Provide your choice (e.g., 0, 1, 2, etc) along with a brief explanation of your reasoning.
Respond only with the following format, nothing else:
Answer: [Provide answer here or 'none of the above']
Rationale: [Provide the rationale here]

Do not include any additional text, headers, or explanations outside this format.
\end{quote}

\noindent\textbf{Self-Critic QA Prompt (GPT-4 Self-Critic):}
\begin{quote}
Q: ... Choices: ... \\
Explain your reasoning. You must choose only one option from above. Your final answer should be a single number (e.g., 0, 1, 2, etc), in the form (answer), at the end of your response. \\
... \\
Review your previous answer and find problems with your answer. \\
... \\
Based on the problems you found, improve your answer. You must choose only one option from above. Please reiterate your answer, with your final answer a single number (e.g., 0, 1, 2, etc), in the form (answer).
\end{quote}

\noindent\textbf{Exploration Prompt (Llama3 DRR):}
\begin{quote}
You are an expert assistant specializing in multiple-choice questions, dedicated to exploring multiple ways of thinking to provide accurate answers. Below, you will see an LLM's previous answer, including the choice it selected and its reasoning, followed by the feedback: 'Wrong answer! Try again.'

Your task is to **think outside the box** and use a **completely different line of reasoning** to approach the question. Carefully reassess each option, explore alternative interpretations, and **avoid repeating the same ideas**. Focus on providing fresh insights and explain your reasoning in a distinct way.

Respond only with the following format, nothing else:
Answer: [Provide answer here or 'none of the above']
Rationale: [Provide the rationale here]

Do not include any additional text, headers, or explanations outside this format.
\end{quote}

\noindent\textbf{Appended Environment Message to Context (Llama3 DRR):}
\begin{quote}
Previous LLM Answer: Answer:... \\
Rationale:... \\
Wrong Answer! Try again.
\end{quote}

\noindent\textbf{Appended Environment Message to Context (GPT-4 DRR):}
\begin{quote}
LLM Answer 1:... \\
Rationale 1:... \\
Env: The previous response was insufficient; explore a new line of reasoning to approach a more accurate answer.
\end{quote}

%%%%%%%%%%%%%%%%%%%%%%%%%%%%%%%%%%%%%%%%%%%%%%%%%%%%%%%%%%%%

\end{document}